\documentclass[12pt]{article}
\usepackage{amsmath,bbm,amssymb}
\usepackage{amsthm}
\usepackage{comment}
\usepackage{graphicx,graphics}
\usepackage{textcomp}
\usepackage{amsfonts}
\usepackage{xcolor}
\usepackage{setspace}
\usepackage{multirow}
\usepackage{arydshln}
\usepackage{booktabs}
\usepackage{longtable}
\usepackage{array}
\usepackage{float}
\usepackage{bm}
\usepackage{mathtools}
\usepackage{enumitem}
\usepackage[numbers,sort&compress]{natbib}
\usepackage[colorlinks=true,linkcolor=blue,citecolor=blue,urlcolor=blue,bookmarks=true,bookmarksnumbered=true,unicode=true]{hyperref}

\newtheorem{assumption}{Assumption}
\newtheorem{theorem}{Theorem}
\newtheorem{lemma}{Lemma}
\newtheorem{definition}{Definition}

\def\v{\boldsymbol}

\DeclareMathOperator{\tr}{tr}

\newcommand{\R}{\mathbf{R}}
\newcommand{\I}{\mathbf{I}}
\newcommand{\J}{\mathbf{J}}
\newcommand{\one}{\boldsymbol{1}}
\newcommand{\y}{\boldsymbol{y}}
\newcommand{\f}{\boldsymbol{f}}
\newcommand{\eps}{\boldsymbol{\varepsilon}}
\newcommand{\Shat}{\mathbf{S}}
\newcommand{\E}{\mathbb{E}}
\newcommand{\Risk}{\mathcal{R}}
\newcommand{\norm}[1]{\lVert #1\rVert}

\begin{document}
\begin{center}
  {\Large\bf Training-point consistency of kernel ridge regression with varying correlation scales}
  \\[2mm]     Weiyan Mu$^{1}$, Shifeng Xiong$^{2,3}$\footnote{Corresponding author, Email: xiong@amss.ac.cn}, Yuming Shen$^{4}$\footnote{Co-Corresponding author, Email: 201310059@gudfs.edu.cn}
		{\footnotesize\\ 1. School of Science, Beijing University of Civil Engineering and Architecture\\ Beijing, China
       \\[1mm] 2. SKLMS, Academy of Mathematics and Systems Science, Chinese Academy of Sciences\\ Beijing, China
      \\[1mm] 3. School of Mathematical Sciences, University of Chinese Academy of Sciences\\ Beijing, China
      \\[1mm] 4. School of Information Science and Technology, Guangdong University of Foreign Studies\\ Guangzhou, China}
\end{center}

\vspace{1cm}\noindent{\bf Abstract}\quad
The scale parameter in kernel ridge regression plays a pivotal role in modeling the correlation structure of data. In modern applications, sampling density and the underlying correlation range often change simultaneously, making a fixed-scale assumption highly restrictive. However, establishing asymptotic theories under varying scale parameters is fundamentally challenging. To address this gap, this paper develops a unified theoretical framework for kernel ridge regression under varying scale parameters. We establish that, on
uniform tensor grids with product exponential kernels, the consistency of the training-point risk holds if and only if the effective local sample size diverges to infinity. This sharp criterion universally covers all regimes along arbitrary deterministic scale sequences, seamlessly unifying the long-range, fixed-scale, and short-range.
Furthermore, we provide an explicit, signal-independent construction of a deterministic regularization penalty that guarantees this consistency. Numerical experiments illustrate the criterion and the corresponding penalty across all the three regimes.

\vspace{2mm}\noindent{{\bf KEY WORDS:} Gaussian process regression; Grid design; Kernel method; Local sample size; Correlation range.}

\section{Introduction}\label{sec:intro}

Consider the nonparametric regression model
\begin{equation}\label{nr}y_i=f(\boldsymbol{x}_i)+\varepsilon_i,\qquad i=1,\ldots,n,\end{equation}
where the design points $\boldsymbol{x}_1,\ldots,\boldsymbol{x}_n\in D\subset\mathbb R^d$, $D$ is a bounded domain, $f:D\to\mathbb R$ is the unknown regression function, and $\varepsilon_1,\ldots,\varepsilon_n$ are random noises.
Kernel ridge regression and Gaussian process regression are fundamental methods for estimating the signal $f$ in nonparametric statistics \cite{wahba1990spline}, machine learning \cite{rasmussen2006gaussian}, spatial statistics \cite{stein1999interpolation}, and computer experiments \cite{baker}.
The two methods combine observations at different sites through a correlation kernel that relies on a range/scale parameter. The scale parameter controls correlation decay: a smaller one makes distant sites more similar, whereas a larger one retains substantial correlation only between near neighbors. Therefore, the estimation performance is highly sensitive to selection of the scale parameter in the kernel \cite{zhang2004inconsistency,kaufman2013range}.

The asymptotic properties of kernel ridge regression and Gaussian process regression have been extensively studied in the literature. Caponnetto and De Vito (2007) established optimal rates for regularized least squares under source conditions \cite{caponnetto2007optimal}. Steinwart, Hush, and Scovel (2009) derived optimal rates for regularized least squares regression \cite{steinwart2009optimal}. Dicker, Foster, and Hsu (2017) clarified the relation between kernel ridge regression and principal component regression through minimax bounds and regularization qualification \cite{dicker2017kernel}. Wang and Jing (2022) derived upper and lower error bounds for Gaussian process regression with possibly misspecified correlation functions and established corresponding convergence results for kernel ridge regression with a deterministic target \cite{wang2022gaussian}. The posterior asymptotics from the Bayesian perspective were investigated in \cite{vandervaart2008rates, vandervaart2009adaptive, vandervaart2011information}. However, these studies are all restricted to a fixed scale parameter. As is known, in many scenarios, particularly when the scale parameter is selected using data-driven methods such as cross-validation, the scale parameter (or its reciprocal) is not necessarily bounded \cite{pramanick}.

Some studies have investigated the case of varying scale parameters for fixed sample sizes \cite{driscoll2002interpolation,barthelme2021spectral,barthelme2023gaussian}. With a diverging sample size, sampling refinement and kernel scale adjustment can occur simultaneously. Therefore, fixed-domain asymptotics should address changing design density and changing correlation range together. To the best of our knowledge, the asymptotic properties of kernel ridge regression and Gaussian process regression with a varying scale parameter have not yet been systematically studied in the literature.

It is fundamentally challenging to extend the asymptotic properties with fixed scale parameters to varying scales. First, as the scale
changes, the kernel matrix undergoes a continuous phase transition from a nearly flat (rank-deficient) structure to an almost diagonal one, rendering classical fixed-kernel spectral tools ineffective. Second, the reproducing kernel Hilbert space (RKHS) norm of a fixed signal depends heavily on the scale, shifting the bias-variance balance dynamically. Third, the simultaneous growth of the sample size and the scale creates a double asymptotic regime governed by the local sample size, where standard empirical process theory struggles to quantify whether local information suffices to offset the noise. Finally, in non-periodic bounded domains, boundary effects are amplified by scale variations and resist Fourier diagonalization. These intertwined challenges necessitate a unified framework that can bridge the long-range, fixed-scale, and short-range regimes.

To overcome these fundamental obstacles, we develop a unified asymptotic framework that fully characterizes the recoverability of signals under varying scale parameters. Our main theorem reveals a sharp phase transition governed entirely by the local sample size: the training-point risk of kernel ridge regression converges to zero if and only if the local sample size goes to infinity. This necessary and sufficient condition is valid for any nonzero signal and any deterministic scale sequence, seamlessly unifying the long-range, fixed-scale, and short-range. Furthermore, we provide a single, signal-independent penalty sequence that realizes this consistency, demonstrating that the bias-variance trade-off can be explicitly balanced across all regimes. The proof circumvents the aforementioned difficulties by combining shape-dependent RKHS norm bounds, an effective dimension estimate via product spectral grouping, and a finite-grid treatment of boundary terms.

The rest of this paper is organized as follows. Section \ref{sec:fixsample} presents the exact risk decomposition and fixed-sample risk with extreme correlation scales. Section \ref{sec:nonperiodic} establishes the unified framework. Section \ref{sec:numerics} reports reproducible numerical experiments. Section \ref{sec:discussion} concludes with discussion. The proofs are given in the Appendix.

\section{Fixed-sample risk with extreme correlation scales}\label{sec:fixsample}

Under the model \eqref{nr}, write $\f_n=(f(\boldsymbol{x}_1),\ldots,f(\boldsymbol{x}_n))'$, abbreviated as $\f$ when the sample size is clear, $\y=(y_1,\ldots,y_n)'$, and $\v{\varepsilon}=(\varepsilon_1,\ldots,\varepsilon_n)'$. Let $K_{\theta}:D\times D\to\mathbb R$ be a positive-semidefinite correlation kernel with a scale parameter $\theta>0$.
Here $\theta$ controls correlation decay: a smaller $\theta$ makes distant sites more similar, whereas a larger $\theta$ retains substantial correlation only between near neighbors. The correlation matrix $\R_\theta$ is defined by $(\R_\theta)_{ij}=K_{\theta}(\boldsymbol{x}_i,\boldsymbol{x}_j)$ for $i,j=1,\ldots,n$.

Kernel ridge regression estimates the regression function $f$ in \eqref{nr} by solving the optimization problem
\begin{equation}\label{penaltyscaling}
 \min_{g\in\mathcal H_{\theta}}
 \left\{\sum_{i=1}^n(y_i-g(\boldsymbol{x}_i))^2
 +\rho\norm{g}_{\mathcal H_{\theta}}^2\right\},
\end{equation}where $\rho>0$ is a tuning penalty parameter, $\mathcal H_{\theta}$ is the reproducing kernel Hilbert space (RKHS) induced by the kernel function $K_{\theta}$, and $\norm{\cdot}_{\mathcal H_{\theta}}$ is the corresponding norm. By \eqref{penaltyscaling}, the fitted training-point vector is
\begin{equation}\label{fv}
 \hat\f_{\theta,\rho}
 =\R_\theta(\R_\theta+\rho\I_n)^{-1}\y
 =:\Shat_{\theta,\rho}\y.
\end{equation}
This formula can also be derived from a Bayesian perspective. Under a zero-mean Gaussian prior with covariance $\tau^2\R_\theta$ and independent Gaussian noise of variance $\sigma^2$, computing the posterior mean at the training points yields \eqref{fv} with $\rho=\sigma^2/\tau^2$.

We study the fixed-design training-point risk
\begin{equation}\label{risk}
 \Risk_n(\theta,\rho;\f)
 =\frac1n\E\norm{\hat\f_{\theta,\rho}-\f}^2.
\end{equation}

\begin{assumption}\label{ass:noise}
The noise satisfies
\begin{equation*}
 \E\eps=0,\qquad \E(\eps\eps')=\sigma^2\I_n,
 \qquad 0<\sigma^2<\infty,
\end{equation*}where $\I_n$ denotes the identity matrix of order $n$.
\end{assumption}

The following theorem presents a bias-variance decomposition of the risk. Let $B_n=B_n(\theta,\rho;\f)$ and $V_n=V_n(\theta,\rho)$ denote the bias and variance of the kernel ridge estimator under the fixed-design training-point risk \eqref{risk}, respectively.
\begin{theorem}\label{th:risk}
Under Assumption~\ref{ass:noise},
\begin{equation}
 \Risk_n(\theta,\rho;\f)
 =\underbrace{\frac{\rho^2}{n}\f'(\R_\theta+\rho\I_n)^{-2}\f}_{=:B_n(\theta,\rho;\f)}
 +\underbrace{\frac{\sigma^2}{n}\tr(\Shat_{\theta,\rho}^2)}_{=:V_n(\theta,\rho)}.
 \label{eq:riskdecomp}
\end{equation}
\end{theorem}

We now study the following two extreme cases\begin{equation*}
 \R_\theta\longrightarrow\J_n:=\one_n\one_n'
 \quad(\theta\downarrow0) \quad\text{and}\quad
 \R_\theta\longrightarrow\I_n
 \quad(\theta\to\infty)
\end{equation*}for fixed $n$ and $\v{f}$, where $\one_n$ denotes the $n$-vector with all elements being one.
Write $\bar f=n^{-1}\one_n'\f$, $\bar y=n^{-1}\one_n'\y$, and $\norm{\f}_n^2=n^{-1}\f'\f$.

\begin{theorem}\label{th:endpoints}
\begin{enumerate}[label=(\roman*),leftmargin=2.2em]
 \item For the long-range correlation case $\R_0=\J_n$,
 \begin{equation*}
  \inf_{\rho>0}\Risk_n(0,\rho;\f)
  =\norm{\f-\bar f\one_n}_n^2+\frac{\bar f^{\,2}\sigma^2}{n\bar f^{\,2}+\sigma^2}.
 \end{equation*}
 \item For the short-range correlation case $\R_\infty=\I_n$, if $\v{f}\neq\v{0}$, then
  \begin{equation*}
  \inf_{\rho>0}\Risk_n(\infty,\rho;\f)
  =\frac{\norm{\f}_n^2\sigma^2}{\norm{\f}_n^2+\sigma^2}.
 \end{equation*}
\end{enumerate}
\end{theorem}

\section{Training-point risk consistency with varying correlation scales}\label{sec:nonperiodic}

Theorem \ref{th:endpoints} shows that, regardless of the choice of $\rho$, the risk does not converge to zero for extreme values of $\theta$ and fixed $n$.
This section considers joint asymptotics as $n$ and $\theta_n$ vary simultaneously, and answers whether there exists $\rho=\rho_n$ such that the risk tends to zero.
For convenience, we denote the sample size by $N$ instead of $n$ in the following.
We introduce the following definition.
\begin{definition}\label{def:consistency}
Given a scale sequence $\{\theta_N\}_{N\geqslant1}$ and a fixed $f$, suppose that there exists a sequence $\{\rho_N\}_{N\geqslant1}$ such that
\begin{equation*}
 \Risk_N(\theta_N,\rho_N;\f_N)\longrightarrow0\quad\text{as}\ N\to\infty.
\end{equation*}
We then say that the corresponding kernel ridge estimator has training-point risk consistency with respect to $\{\theta_N\}_{N\geqslant1}$ for $f$.
\end{definition}

Because the risk is the second moment of the squared training-point error, its convergence to zero is equivalent to convergence of the training-point error in the corresponding  $L_2$ space. Markov's inequality further yields convergence in probability.

In our analysis, the dimension $d$ is fixed.
We need the following assumptions on the design and the correlation function.

\begin{assumption}\label{ass:design}
Let $D=[0,1]^d$ and $N=n^d$ with some integer $n$. The design points are
\begin{equation*}
 \boldsymbol{x}_{\boldsymbol i}=\frac{\boldsymbol i}{n},\qquad
 \boldsymbol i=(i_1,\ldots,i_d)\in\{1,\ldots,n\}^d.
\end{equation*}
\end{assumption}

\begin{assumption}\label{ass:kernel}
For $\theta>0$, the correlation function is
\begin{equation*}
 K_{\theta}(\boldsymbol{x},\boldsymbol{y})
 =\exp\{-\theta\|\boldsymbol{x}-\boldsymbol{y}\|_1\}
 =\prod_{r=1}^d e^{-\theta|x_r-y_r|},
 \qquad \boldsymbol{x},\boldsymbol{y}\in[0,1]^d.
\end{equation*}
\end{assumption}

Under Assumptions \ref{ass:design} and \ref{ass:kernel}, we denote the correlation matrix by $\R^{(d)}=\R^{(d)}(\theta)$ with
\begin{equation}\label{R}
 \R^{(d)}_{\boldsymbol i,\boldsymbol j}(\theta)=K_{\theta}(\boldsymbol{x}_{\boldsymbol i},\boldsymbol{x}_{\boldsymbol j})
 =q^{\|\boldsymbol i-\boldsymbol j\|_1}
 =\prod_{r=1}^d q^{|i_r-j_r|}
\end{equation}for $\boldsymbol i=(i_1,\ldots,i_d), \boldsymbol j=(j_1,\ldots,j_d)\in\{1,\ldots,n\}^d$, where $q=e^{-\theta/n}$.
Thus \begin{equation}\label{R1d}\R^{(d)}=\R_{\mathrm{1d}}(\theta)^{\otimes d}\quad \text{with}\quad (\R_{\mathrm{1d}}(\theta))_{ij}=q^{|i-j|},\quad i,j=1,\ldots,n.\end{equation} Write
\begin{equation*}
 m_N=\frac{n}{\theta_N},\qquad M_N=m_N^d=\frac{N}{\theta_N^d},
\end{equation*}which can be viewed as local sample sizes in one dimension and $d$ dimension, respectively.

For a multi-index $\boldsymbol{\alpha}=(\alpha_1,\ldots,\alpha_d)'\in\{0,1\}^d$, write $\partial^{\boldsymbol{\alpha}}=\partial_1^{{\alpha}_1}\cdots\partial_d^{{\alpha}_d}$. Let
\begin{equation*}
 H^{1,\ldots,1}([0,1]^d)
 =\left\{f:\partial^{\boldsymbol{\alpha}} f\in L_2([0,1]^d),
 \ \boldsymbol{\alpha}\in\{0,1\}^d\right\},
\end{equation*}denote the mixed first-order Sobolev space equipped with the norm
\begin{equation}\label{snorm}\|f\|_{1,\mathrm{mix}}^2
=\sum_{\boldsymbol{\alpha}\in\{0,1\}^d}\|\partial^{\boldsymbol{\alpha}} f\|_{L_2}^2.\end{equation}
This space is the reproducing-kernel Hilbert space of the product exponential kernel at a fixed scale, and its elements have continuous representatives. We always take the continuous representative, and grid evaluations and pointwise Riemann sums are well defined.

The following theorem presents the necessary and sufficient condition for training-point risk consistency.
\begin{theorem}\label{thm:main}
Under Assumptions~\ref{ass:noise}, \ref{ass:design}, and \ref{ass:kernel}, let $f\in H^{1,\ldots,1}([0,1]^d)$ be fixed and nonzero, and let $\{\theta_N\}_{N\geqslant1}$ be an arbitrary deterministic sequence of positive numbers. The corresponding kernel ridge estimator has training-point risk consistency with respect to $\{\theta_N\}_{N\geqslant1}$ for $f$ if and only if
\begin{equation}
 M_N=\frac{N}{\theta_N^d}\longrightarrow\infty.
 \label{eq:maincriterion}
\end{equation}
More specifically, the following two statements hold.
\begin{enumerate}[label=(\roman*),leftmargin=2.2em]
 \item If $M_N\to\infty$, selecting
 \begin{equation}
  \rho_N=
  \begin{cases}
   N\theta_N^d/\log N,&\theta_N\leqslant1,\\[2pt]
   M_N^{2/3},&\theta_N>1,
  \end{cases}
  \label{eq:unifiedrho}
 \end{equation}
 gives $\Risk_N(\theta_N,\rho_N;\f_N)\to0$.
 \item If $\liminf_NM_N<\infty$, then for every deterministic sequence $\rho_N>0$,
 \begin{equation}
  \limsup_{N\to\infty}\Risk_N(\theta_N,\rho_N;\f_N)>0 .
  \label{eq:positiverisk}
 \end{equation}
\end{enumerate}
\end{theorem}

This theorem covers all the cases of $\theta_N\to0$, fixed or bounded $\theta_N$, and $\theta_N\to\infty$.
The two branches of \eqref{eq:unifiedrho} have different orders at $\theta_N=1$. This discontinuity arises from using the long-range spectral estimate and the coarse short-range variance bound on different sides; it serves only the consistency proof and does not represent a discontinuity in an optimal penalty or an optimal risk rate.

In the classical theory on consistency of kernel ridge regression for fixed $\theta$, it is required that $\rho_N\to\infty$ and $\rho_N/N\to0$; see, e.g. \cite{caponnetto2007optimal}. Set $\theta_N$ to be fixed in Theorem \ref{thm:main}, and our $\rho_N$ in \eqref{eq:unifiedrho} satisfies this condition.

Sanz-Alonso (2026) gave a scale-consistent risk analysis of Mat\'ern kernel ridge regression on periodic domains with quasi-uniform designs and Sobolev target classes \cite{sanz2026matern}. Its kernel-scale information quantity has the same order as our local effective sample size at fixed dimension and fixed signal-to-noise ratio.
Wenzel and Santin (2026) studied asymptotically optimal shape parameters for finitely smooth Sobolev kernels and their relation to superconvergence \cite{wenzel2026shape}. Their results concern asymptotic approximation rates for noiseless kernel interpolation, whereas we study noisy kernel ridge regression.

\section{Numerical experiments}\label{sec:numerics}

We now illustrate the theoretical results with numerical examples. Consider the following one-dimensional and two-dimensional test functions,
\begin{align*}
 f_1(x)&=1+0.4\cos(2\pi x)+0.2x,\\
 f_2(x_1,x_2)&=1+0.3\cos(2\pi x_1)+0.2\sin(2\pi x_2)+0.1x_1x_2,
\end{align*} in the regression model \eqref{nr}. For the one-dimensional case, the sample size is set as $n\in\{64,125,216,343,512,729,1000\}$. For the two-dimensional case, set $n\in\{16, 20, 24, 28, 32, 36, 40\}$ and $N =n^2$, Let $\sigma^2=1$ in Assumption \ref{ass:noise}. Under the grid design in Assumption \ref{ass:design}, generate data with normal noises, and compute the kernel ridge estimator with the correlation function in Assumption \ref{ass:kernel}.

For the one-dimensional case, consider the following four cases of the correlation parameter $\theta$ in kernel ridge regression: (i) fixed $\theta=1/2$, corresponding to the local sample size $M_n=2n$; (ii) short range $\theta_n=n^{2/3}$, corresponding to $M_n=n^{1/3}$; (iii) long range $\theta_n=n^{-3/2}$, corresponding to $M_n=n^{5/2}$; (iv) too short range $\theta_n=n/4$, corresponding to bounded local sample size $M_n=4$. Cases (i)-(iii) satisfy the diverging-local-sample-size condition \eqref{eq:maincriterion} in Theorem \ref{thm:main}, and we use $\rho_n$ in \eqref{eq:unifiedrho}. For case (iv) where condition \eqref{eq:maincriterion} does not hold, we select $\rho$ that yields the minimum value of training-point risk.

For the two-dimensional case, consider the following four cases similarly: (i) fixed $\theta=1/2$, corresponding to the local sample size $M_N=4N$; (ii) short range $\theta_N=N^{1/4}$, corresponding to $M_N=N^{1/2}$; (iii) long range $\theta_N=N^{-1}$, corresponding to $M_N=N^3$; (iv) too short range $\theta_N=N^{1/2}/4$, corresponding to bounded local sample size $M_N=16$. Cases (i)-(iii) satisfy condition \eqref{eq:maincriterion}, while case (iv) does not. We select $\rho$ in the same manner as in the one-dimensional case.

\begin{figure}[t]
 \centering
 \includegraphics[width=\textwidth]{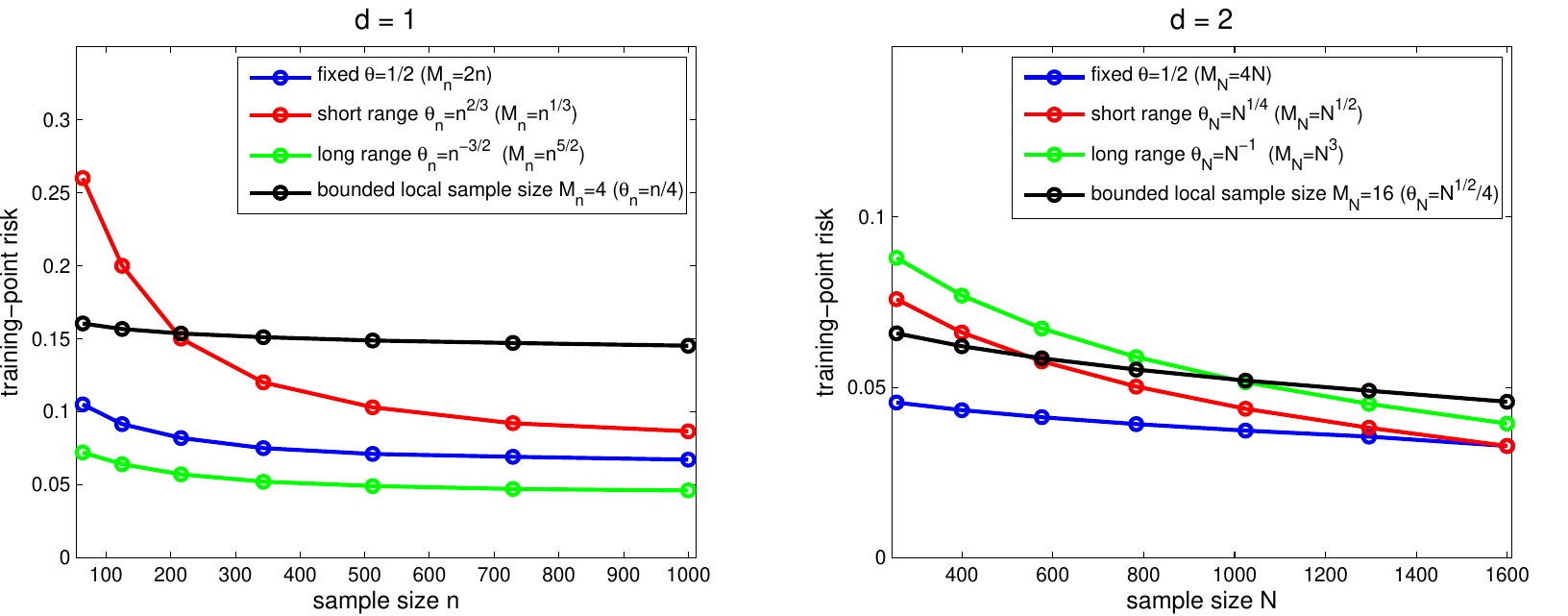}
 \caption{Training-point risk under four correlation regimes in one and two dimensions. }
 \label{fig:risk}
\end{figure}

We compute the fixed-design training-point risk \eqref{risk} by the analytic formula \eqref{eq:riskdecomp}, not a Monte Carlo average over random realizations. The results are displayed in Figure \ref{fig:risk}. It can be seen that the risk converges to zero, albeit slowly, if the diverging-local-sample-size condition holds, and remains bounded away from zero otherwise. The performance is consistent with the sufficiency and the necessity of the diverging-local-sample-size condition. The slow convergence may be due to the fact that we provide only a condition for convergence, rather than a condition for achieving the optimal convergence rate.

\section{Discussion}\label{sec:discussion}

This paper establishes a unified signal-recovery framework on fixed-dimensional bounded uniform grids with product exponential kernels, placing long-range correlation, fixed scale, and short-range correlation under one necessary and sufficient criterion and one penalty construction. For every fixed nonzero signal in the mixed first-order Sobolev space, training-point risk consistency is equivalent to the diverging-local-sample-size condition $M_N=N/\theta_N^d\to\infty$. The statement holds along the entire sequence and therefore also covers mixed scale paths containing subsequences from different regimes.

The present study is restricted to specific conditions, leaving several open problems for future research. First, it is unclear whether the diverging-local-sample-size condition
is valid for general Mat\'ern kernels, nonuniform designs, out-of-sample prediction, and continuous $L_2$ recovery. Second, we establish the validity of prespecified deterministic penalty sequences, while the conditions under which cross-validation and other data-driven rules succeed are left open. Third, the numerical results show that the penalty sequence we proposed is far from optimal, and regime-specific optimal penalties and optimal risk rates remain to be studied.

\section*{Appendix: Proofs}

\subsection*{Proofs in Section \ref{sec:fixsample}}

\noindent\emph{Proof of Theorem \ref{th:risk}}. Right-multiplying $(\R_\theta+\rho\I_n)-\R_\theta=\rho\I_n$ by $(\R_\theta+\rho\I_n)^{-1}$ gives
\begin{equation*}
 \I_n-\Shat_{\theta,\rho}
 =\rho(\R_\theta+\rho\I_n)^{-1}.
\end{equation*}
Consequently,
\begin{equation*}
 \hat\f_{\theta,\rho}-\f
 =-\rho(\R_\theta+\rho\I_n)^{-1}\f+\Shat_{\theta,\rho}\eps.
\end{equation*}
Expanding the squared norm, we find that the expected cross term is
\begin{equation*}
 \E\left[\eps'\Shat_{\theta,\rho}'
 (\Shat_{\theta,\rho}-\I_n)\f\right]=0,
\end{equation*}
because $\E\eps=0$. The deterministic part divided by $n$ is $B_n$. For the noise part,
\begin{align*}
 \E\norm{\Shat_{\theta,\rho}\eps}^2
 &=\E\tr\left(\Shat_{\theta,\rho}\eps\eps'\Shat_{\theta,\rho}'\right)\\
 &=\tr\left(\Shat_{\theta,\rho}\E(\eps\eps')\Shat_{\theta,\rho}'\right)\\
 &=\sigma^2\tr(\Shat_{\theta,\rho}^2),
\end{align*}
where the last step uses the symmetry of $\Shat_{\theta,\rho}$. Division by $n$ gives \eqref{eq:riskdecomp}.
\qed

\noindent\emph{Proof of Theorem \ref{th:endpoints}}. The identity $\J_n^2=n\J_n$ directly gives
\begin{equation*}
 (\J_n+\rho\I_n)^{-1}
 =\frac1\rho\I_n-\frac1{\rho(n+\rho)}\J_n,
 \qquad
 \J_n(\J_n+\rho\I_n)^{-1}=\frac{\J_n}{n+\rho}.
\end{equation*}
This yields the fitted vector $\hat\f_{0,\rho}={n\bar y\one_n}/(n+\rho)$. Orthogonally decomposing $\f$ into $\bar f\one_n$ and $\f-\bar f\one_n$, and using $\tr(\J_n^2)=n^2$, gives $$\Risk_n(0,\rho;\f)
  =\norm{\f-\bar f\one_n}_n^2
  +\frac{\rho^2\bar f^{\,2}}{(n+\rho)^2}
  +\frac{n\sigma^2}{(n+\rho)^2},$$which reaches its minimum at $\rho=\sigma^2/\bar f^{\,2}$, and this completes the proof of (i).

By $\R_\infty=\I_n$, we can immediately get $$\hat\f_{\infty,\rho}=\frac{\y}{1+\rho},
  \qquad
  \Risk_n(\infty,\rho;\f)
  =\frac{\rho^2\norm{\f}_n^2+\sigma^2}{(1+\rho)^2}.$$ Some algebra gives its minimizer $\rho=\sigma^2/\norm{\f}_n^2$. Substitution gives (ii).
\qed

\subsection*{Proofs in Section \ref{sec:nonperiodic}}

For an $N\times N$ correlation matrix $\R$ and $N$-vector $\v{a}$, define $\|\v{a}\|_{\R^{-1}}^2=\v{a}'\R^{-1}\v{a}$, and let $\|\R\|_{\mathrm{op}}$ denote the operator 2-norm/maximum singular value of $\R$. Let $\mathcal H_{\theta}$ denote the reproducing-kernel Hilbert space of the continuous product exponential kernel in Assumption~\ref{ass:kernel} equipped with the norm $\|\cdot\|_{H_{\theta}}$. The following lemma presents the general risk bounds of kernel ridge regression. As in Theorem \ref{th:risk}, $B_N$ and $V_N$ denote the bias and variance of the kernel ridge estimator under the fixed-design training-point risk \eqref{risk}, respectively.

\begin{lemma}\label{lem:generalbounds}
Under Assumption \ref{ass:noise}, we have
\begin{align}
 &\text{the upper bounds}\quad B_N\leqslant\frac{\rho}{4N}\|\f\|_{\R^{-1}}^2,
 \qquad V_N\leqslant\frac{\sigma^2}{4\rho};
 \label{eq:upperBV}\\&\text{and the lower bounds}\quad
 B_N\geqslant\frac{\|\f\|_N^2}{(1+\|\R\|_{\mathrm{op}}/\rho)^2},
 \qquad V_N\geqslant\frac{\sigma^2}{(\|\R\|_{\mathrm{op}}+\rho)^2}.
 \label{eq:lowerBV}
\end{align}
Moreover, under Assumptions \ref{ass:noise}, \ref{ass:design}, and \ref{ass:kernel}, if $\R=\R^{(d)}$ and $f\in H^{1,\ldots,1}$, then
\begin{equation}
 \|\f\|_{\R^{-1}}^2
 \leqslant \|f\|_{\mathcal H_{\theta}}^2 .
 \label{eq:restrictionnorm}
\end{equation}
\end{lemma}

\begin{proof}
Write the eigenvalue decomposition $\R=\mathbf{U}\operatorname{diag}(\lambda_1,\ldots,\lambda_N)\mathbf{U}'$. By the reproducing-kernel representation, there exists $\boldsymbol{g}$ such that $\f=\R^{1/2}\boldsymbol{g}$ and $\|\boldsymbol{g}\|_2=\|\f\|_{\R^{-1}}$. Therefore,
\begin{align*}
 B_N
 &=\frac{\rho^2}{N}\boldsymbol{g}'\R^{1/2}(\R+\rho\I)^{-2}\R^{1/2}\boldsymbol{g}\\
 &\leqslant\frac{\|\boldsymbol{g}\|_2^2}{N}
 \sup_{\lambda\geqslant0}\frac{\rho^2\lambda}{(\lambda+\rho)^2}
 =\frac{\rho}{4N}\|\f\|_{\R^{-1}}^2 .
\end{align*}
Since $4\rho\lambda\leqslant(\lambda+\rho)^2$ and $\sum_j\lambda_j=\tr(\R)=N$,
\begin{equation*}
 V_N=\frac{\sigma^2}{N}
 \sum_{j=1}^N\frac{\lambda_j^2}{(\lambda_j+\rho)^2}
 \leqslant\frac{\sigma^2}{4N\rho}\sum_{j=1}^N\lambda_j
 =\frac{\sigma^2}{4\rho}.
\end{equation*}

For the lower bounds, $0\leqslant\lambda_j\leqslant \|\R\|_{\mathrm{op}}$ gives
\begin{equation*}
 B_N\geqslant\frac{\rho^2}{N(\|\R\|_{\mathrm{op}}+\rho)^2}\f'\f
 =\frac{\|\f\|_N^2}{(1+\|\R\|_{\mathrm{op}}/\rho)^2}.
\end{equation*}
Similarly, using $\|\R\|_F^2\geqslant\sum_{i=1}^N\R_{ii}^2=N$ with the Frobenius norm $\|\cdot\|_F$, we obtain
\begin{align*}
 V_N
 &\geqslant\frac{\sigma^2}{N(\|\R\|_{\mathrm{op}}+\rho)^2}
 \sum_{j=1}^N\lambda_j^2
 =\frac{\sigma^2\|\R\|_F^2}{N(\|\R\|_{\mathrm{op}}+\rho)^2}
 \geqslant\frac{\sigma^2}{(\|\R\|_{\mathrm{op}}+\rho)^2}.
\end{align*}
Finally, minimum-norm interpolation \cite{Wendland} gives
\begin{equation*}
 \|\f\|_{\R^{-1}}^2
 =\min\{\|g\|_{\mathcal H_{\theta}}^2:
 g(\boldsymbol{x}_{\boldsymbol i})=f(\boldsymbol{x}_{\boldsymbol i}),\ \boldsymbol i\in\{1,\ldots,n\}^d\},
\end{equation*}
and taking $g=f$ yields \eqref{eq:restrictionnorm}.
\end{proof}

The following lemma shows the effect of the correlation scale on the RKHS norm.
\begin{lemma}\label{lem:rkhsnorm}Under Assumptions \ref{ass:design} and \ref{ass:kernel},
for $f\in H^{1,\ldots,1}([0,1]^d)$, there exists a constant $C_f$, depending only on $d$ and $f$, such that
\begin{equation}
 \|f\|_{\mathcal H_{\theta}}^2\leqslant
 \begin{cases}
  C_f\theta^{-d},&0<\theta\leqslant1,\\[2pt]
  C_f\theta^d,&\theta\geqslant1 .
 \end{cases}
 \label{eq:rkhsshape}
\end{equation}
\end{lemma}

\begin{proof}
We first construct a bounded extension of $f$ to $\mathbb R^d$. Taking the even reflection of $f$ successively about $0$ and $1$ in each coordinate gives an extension $f_E$ on $[-1,2]^d$. Each even reflection preserves the first weak derivative in that coordinate and its $L_2$ integrability. Coordinatewise application therefore preserves all mixed weak derivatives and gives a Sobolev-norm bound depending only on $d$.

Next, take an infinitely differentiable function $\eta$, defined on $\mathbb{R}$, which equals $1$ on $[0,1]$ and is supported in $(-1/2,3/2)$. Let $\Phi(\boldsymbol{x})=\prod_{r=1}^d\eta(x_r)$ and $\widetilde f=\Phi f_E$. Then $\widetilde f$ agrees with $f$ on $[0,1]^d$ and has compact support. Applying the multivariate Leibniz rule to compute the partial derivatives term by term for $\boldsymbol{\alpha}\in\{0,1\}^d$ gives
\begin{equation}\label{eq:extension}
 \sum_{\boldsymbol{\alpha}\in\{0,1\}^d}
 \|\partial^{\boldsymbol{\alpha}}\widetilde f\|_{L_2(\mathbb R^d)}^2
 \leqslant C_d\|f\|_{1,\mathrm{mix}}^2,
\end{equation}where $C_d$ is a constant depending only on $d$ and the norm $\|\cdot\|_{1,\mathrm{mix}}$ is defined by \eqref{snorm}. The right-hand side of \eqref{eq:extension} is therefore a constant that depends only on $d$ and $f$.
Regard the kernel function $K_{\theta}$ on $[0,1]^d$ as the restriction of a kernel on $\mathbb R^d$. Let $\mathcal H_{\theta}(\mathbb{R}^d)$ denote the RKHS induced by this kernel on $\mathbb R^d$. The original RKHS $\mathcal H_{\theta}$ induced by the restricted kernel consists of restrictions of functions in $\mathcal H_{\theta}(\mathbb{R}^d)$. We have \begin{equation}\label{norme}\|f\|_{\mathcal H_{\theta}}\leqslant\|\widetilde f\|_{\mathcal H_{\theta}(\mathbb R^d)}.\end{equation}

Consider the Fourier transform
$\widehat g(\boldsymbol{\omega})=\int_{\mathbb R^d}g(\boldsymbol{x})e^{-i\boldsymbol{\omega}'\boldsymbol{x}}\,d\boldsymbol{x}$.
The spectral density of the one-dimensional exponential kernel is
\begin{equation*}
 s_\theta(\omega)=\int_{-\infty}^{\infty}
 e^{-\theta|t|}e^{-i\omega t}\,dt
 =\frac{2\theta}{\theta^2+\omega^2},
\end{equation*}
and thus the product kernel has spectral density $\prod_{r=1}^ds_\theta(\omega_r)$. The Fourier representation of the RKHS norm and Parseval's identity \cite{stein1999interpolation,Wendland} give
\begin{align*}
 \|\widetilde f\|_{\mathcal H_{\theta,d}(\mathbb R^d)}^2
 &=\frac1{(2\pi)^d}\int_{\mathbb R^d}
 |\widehat{\widetilde f}(\boldsymbol{\omega})|^2
 \prod_{r=1}^d\frac{\theta^2+\omega_r^2}{2\theta}\,d\boldsymbol{\omega}\\
 &=2^{-d}\sum_{\boldsymbol{\alpha}\in\{0,1\}^d}
 \theta^{d-2|\boldsymbol{\alpha}|}
 \|\partial^{\boldsymbol{\alpha}}\widetilde f\|_{L_2(\mathbb R^d)}^2 .
\end{align*}
If $\theta\geqslant1$, then $\theta^{d-2|\boldsymbol{\alpha}|}\leqslant\theta^d$; if $0<\theta\leqslant1$, then $\theta^{d-2|\boldsymbol{\alpha}|}\leqslant\theta^{-d}$. Combining \eqref{eq:extension} and \eqref{norme} proves \eqref{eq:rkhsshape}.
\end{proof}

Lemma~\ref{lem:rkhsnorm} unifies the two bias controls. For $\theta\geqslant1$, \eqref{eq:upperBV} and \eqref{eq:rkhsshape} give
\begin{equation}
 B_N\leqslant C_f\frac{\rho\theta^d}{N}
 =C_f\frac{\rho}{M_N};
 \label{eq:biaslarge}
\end{equation}
for $0<\theta\leqslant1$, they give
\begin{equation}
 B_N\leqslant C_f\frac{\rho}{N\theta^d}.
 \label{eq:biassmall}
\end{equation}
The constant $C_f$ may differ between the two cases.

The coarse variance bound $V_N\leqslant\sigma^2/(4\rho)$ in \eqref{eq:upperBV} remains valid under extremely long correlation, but a sharper spectral bound will be presented. We first estimate the nonprincipal eigenvalues of the one-dimensional factor in the following lemma.

\begin{lemma}\label{lem:1deigen}
Under Assumptions \ref{ass:design} and \ref{ass:kernel}, let $\Lambda_{n,0}\geqslant\Lambda_{n,1}\geqslant\cdots\geqslant\Lambda_{n,n-1}$ be the eigenvalues of $\R_{\mathrm{1d}}(\theta)$ in \eqref{R1d}. There exists an absolute constant $C_0>0$ such that, for all $n\geqslant2$, $0<\theta\leqslant1$, and $1\leqslant j\leqslant n-1$,
\begin{equation}
 \Lambda_{n,j}\leqslant C_0\frac{n\theta}{j^2}.
 \label{eq:1deigenbound}
\end{equation}
\end{lemma}

\begin{proof}
Note that $q=e^{-\theta/n}$ in \eqref{R}. We have
\begin{equation*}
 \R_{\mathrm{1d}}(\theta)^{-1}=\frac1{1-q^2}
 \begin{pmatrix}
 1&-q&&&\\
 -q&1+q^2&-q&&\\
 &\ddots&\ddots&\ddots&\\
 &&-q&1+q^2&-q\\
 &&&-q&1
 \end{pmatrix}.
\end{equation*}
Direct expansion for every $\boldsymbol{v}=(v_1,\ldots,v_n)'\in\mathbb R^n$ gives
\begin{align*}
 (1-q^2)\boldsymbol{v}'\R_{\mathrm{1d}}(\theta)^{-1}\boldsymbol{v}
 ={}&q\sum_{i=1}^{n-1}(v_i-v_{i+1})^2
 +(1-q)(v_1^2+v_n^2)\nonumber\\
 &+(1-q)^2\sum_{i=2}^{n-1}v_i^2.
\end{align*}
Discarding the last two nonnegative terms, and letting $\mathbf{L}_n$ be the path-graph Laplacian \cite{Chung} defined by $\boldsymbol{v}'\mathbf{L}_n\boldsymbol{v}=\sum_{i=1}^{n-1}(v_i-v_{i+1})^2$, yields
\begin{equation*}
 \R_{\mathrm{1d}}(\theta)^{-1}\succeq
 \frac{q}{1-q^2}\mathbf{L}_n
 =\frac1{2\sinh(\theta/n)}\mathbf{L}_n.
\end{equation*}
Since $0<\theta/n\leqslant1$ and $\sinh t\leqslant(\sinh1)t$ for $0<t\leqslant1$, taking $t=\theta/n$ gives
\begin{equation*}
 \R_{\mathrm{1d}}(\theta)^{-1}\succeq\frac1{2\sinh1}\frac n\theta \mathbf{L}_n,
 \qquad c_1:=\frac1{2\sinh1}.
\end{equation*}

The $j$th eigenvalue of the path-graph Laplacian satisfies
\begin{equation*}
 \nu_{n,j}=4\sin^2\frac{\pi j}{2n}
 \geqslant\frac{4j^2}{n^2},
 \qquad 1\leqslant j\leqslant n-1,
\end{equation*}
where we use $\sin t\geqslant2t/\pi$ for $0\leqslant t\leqslant\pi/2$. Applying the Courant--Fischer theorem \cite{Horn} to the increasingly ordered eigenvalues of $\R_{\mathrm{1d}}(\theta)^{-1}$ gives
\begin{equation*}
 \Lambda_{n,j}^{-1}
 \geqslant c_1\frac{n}{\theta}\nu_{n,j}
 \geqslant 4c_1\frac{j^2}{n\theta}.
\end{equation*}
Taking reciprocals proves \eqref{eq:1deigenbound}.
\end{proof}

The following lemma provides a sharper upper bound on the variance in Lemma \ref{lem:generalbounds} under long correlation.

\begin{lemma}\label{lem:longvariance}
Under Assumptions \ref{ass:noise}, \ref{ass:design}, and \ref{ass:kernel}, let $0<\theta\leqslant1$, $\R=\R^{(d)}$, and $A=N\theta^d/\rho$. For $A\geqslant1$, there exists a constant $C_d$, depending only on $d$, such that
\begin{equation}
 V_N\leqslant C_d\sigma^2
 \left\{n^{-1}+\frac{A^{1/2}(1+\log A)^{d-1}}{N}\right\}.
 \label{eq:longvariance}
\end{equation}
\end{lemma}

\begin{proof}
The eigenvalues of the product matrix $\R^{(d)}=\R_{\mathrm{1d}}(\theta)^{\otimes d}$ are indexed by multi-indices $\boldsymbol j=(j_1,\ldots,j_d)\in\{0,\ldots,n-1\}^d$:
\begin{equation*}
 \Lambda_{\boldsymbol j}=\prod_{r=1}^d\Lambda_{n,j_r}.
\end{equation*}
By Theorem \ref{th:risk}, $$V_N=\frac{\sigma^2}{N}\tr(\Shat_{\theta,\rho}^2)=\frac{\sigma^2}{N}\sum_{\boldsymbol j}\left(\frac{\Lambda_{\boldsymbol j}}{\Lambda_{\boldsymbol j}+\rho}\right)^2.$$
For $t\geqslant0$, the inequalities $t^2/(1+t)^2\leqslant t/(1+t)\leqslant\min\{1,t\}$ give
\begin{equation}\label{v}
 V_N\leqslant\frac{\sigma^2}{N}
 \sum_{\boldsymbol j}\min\left\{1,\frac{\Lambda_{\boldsymbol j}}{\rho}\right\}.
\end{equation}
We group the terms by $k=\#\{r=1,\ldots,d: j_r\geqslant1\}$.

When $k<d$, the group contains at most $\binom dkn^k$ multi-indices. Bounding each summand by $1$ and adding the groups $k=0,\ldots,d-1$, we find that their contribution to $V_N$ is at most
\begin{equation}
 \frac{\sigma^2}{N}
 \sum_{k=0}^{d-1}\binom dkn^k
 \leqslant C_d\sigma^2n^{-1}.
 \label{eq:lowgroups}
\end{equation}

When $k=d$, every $j_r\geqslant1$. Lemma \ref{lem:1deigen} gives
\begin{equation*}
 \frac{\Lambda_{\boldsymbol j}}{\rho}
 \leqslant\frac{C_0^dn^d\theta^d}{\rho\prod_{r=1}^dj_r^2}
 =\frac{C_0^dA}{\prod_{r=1}^dj_r^2}.
\end{equation*}
Consequently, the sum over finite terms in the right-hand side of \eqref{v} is controlled by the unrestricted sum, i.e.,
\begin{equation*}
 \sum_{\boldsymbol j,\ k=d}\min\left\{1,\frac{\Lambda_{\boldsymbol j}}{\rho}\right\}\leqslant\sum_{\boldsymbol j\in\mathbb N^d}\min\left\{1,\frac{\Lambda_{\boldsymbol j}}{\rho}\right\}\leqslant\mathcal S_d(B):=\sum_{\boldsymbol j\in\mathbb N^d}
 \min\left\{1,\frac{B}{\prod_{r=1}^dj_r^2}\right\},
\end{equation*}where $B=C_0^dA\geqslant1$ (we may take $C_0\geqslant1$ in Lemma \ref{lem:1deigen}, and thus $A\geqslant1$ implies $B\geqslant1$).

We now estimate both the body region $\prod_rj_r\leqslant B^{1/2}$ and all remaining tail terms.
Define the lattice count
\begin{equation*}
 \mathcal Q_d(T):=
 \#\left\{\boldsymbol j\in\mathbb N^d:
 \prod_{r=1}^dj_r\leqslant T\right\}.
\end{equation*}
Clearly, $\mathcal Q_1(T)\leqslant T$. If $\mathcal Q_{d-1}(T)\leqslant C T(1+\log T)^{d-2}$, then
\begin{align}
 \mathcal Q_d(T)
 &=\sum_{\substack{j\in\mathbb N\\j\leqslant T}}\mathcal Q_{d-1}(T/j)\nonumber\\
 &\leqslant C T\sum_{j\leqslant T}j^{-1}
 \left(1+\log\frac{T}{j}\right)^{d-2}
 \leqslant C'T(1+\log T)^{d-1}.
 \label{eq:latticeinduction}
\end{align}
Hence the body-region contribution is at most
\begin{equation}
 \mathcal Q_d(B^{1/2})
 \leqslant C_dB^{1/2}(1+\log B)^{d-1}.
 \label{eq:bodyregion}
\end{equation}

For the tail, let $\ell_0=\lfloor\log_2(B^{1/2})\rfloor+1$ and use the dyadic shells
\begin{equation*}
 \mathcal D_\ell=\left\{\boldsymbol j\in\mathbb N^d:
 2^{\ell-1}<\prod_{r=1}^dj_r\leqslant2^\ell\right\},
 \qquad \ell\geqslant\ell_0 .
\end{equation*}
These shells cover every index with $\prod_rj_r>B^{1/2}$. Any overlap with the body region is negligible since our estimate is used for an upper bound. Every term in $\mathcal D_\ell$ satisfies
\begin{equation*}
 \min\left\{1,\frac{B}{\prod_rj_r^2}\right\}
 \leqslant\frac{B}{2^{2\ell-2}},
 \qquad
 |\mathcal D_\ell|\leqslant\mathcal Q_d(2^\ell).
\end{equation*}
Combining this with \eqref{eq:latticeinduction}, we bound the tail contribution by
\begin{align}
 \sum_{\ell=\ell_0}^{\infty}
 \frac{B}{2^{2\ell-2}}\mathcal Q_d(2^\ell)
 &\leqslant C_dB\sum_{\ell=\ell_0}^{\infty}
 2^{-\ell}(1+\ell)^{d-1}\nonumber\\
 &\leqslant C_dB2^{-\ell_0}(1+\ell_0)^{d-1}\nonumber\\
 &\leqslant C_dB^{1/2}(1+\log B)^{d-1}.
 \label{eq:tailregion}
\end{align}
The second inequality uses the ratio of consecutive terms, $2^{-1}((\ell+2)/(\ell+1))^{d-1}\to1/2$: beyond a sufficiently large $\ell$, comparison with a geometric series applies, and the finitely many earlier terms are absorbed into a constant depending only on $d$.

Combining \eqref{eq:bodyregion} and \eqref{eq:tailregion} gives
\begin{equation*}
 \mathcal S_d(B)\leqslant C_dB^{1/2}(1+\log B)^{d-1}.
\end{equation*}
Substituting $B=C_0^dA$ and absorbing constants depending only on $d$, we find that the full-coordinate group contributes at most
\begin{equation}
 C_d\sigma^2
 \frac{A^{1/2}(1+\log A)^{d-1}}{N}
 \label{eq:fullgroup}
\end{equation}
to $V_N$. Combining \eqref{eq:lowgroups} and \eqref{eq:fullgroup} implies \eqref{eq:longvariance}.
\end{proof}

\begin{proof}[Proof of Theorem~\ref{thm:main}]
(i) We first prove sufficiency. Suppose $M_N\to\infty$.

When $0<\theta_N\leqslant1$, take
\begin{equation*}
 A_N=\log N,\qquad \rho_N=\frac{N\theta_N^d}{A_N}.
\end{equation*}
By \eqref{eq:restrictionnorm}, Lemma~\ref{lem:rkhsnorm}, and \eqref{eq:biassmall},
\begin{equation*}
 B_N\leqslant C_f\frac{\rho_N}{N\theta_N^d}
 =\frac{C_f}{A_N}\longrightarrow0 .
\end{equation*}
For sufficiently large $N$, $A_N\geqslant1$. Lemma \ref{lem:longvariance} therefore gives
\begin{equation*}
 V_N\leqslant C_d\sigma^2
 \left\{n^{-1}+
 \frac{(\log N)^{1/2}(1+\log\log N)^{d-1}}{N}\right\}
 \longrightarrow0 .
\end{equation*}
Thus the risk tends to zero in this regime.

When $\theta_N>1$, take $\rho_N=M_N^{2/3}$. By \eqref{eq:biaslarge} and \eqref{eq:upperBV},
\begin{equation*}
 B_N\leqslant C_f\frac{\rho_N}{M_N}=C_fM_N^{-1/3},
 \qquad
 V_N\leqslant\frac{\sigma^2}{4}M_N^{-2/3}.
\end{equation*}
Both terms tend to zero as $M_N\to\infty$. Combining the above two cases proves the sufficiency of \eqref{eq:unifiedrho}.

(ii) We now prove necessity. Suppose $\liminf_NM_N<\infty$. Choose a subsequence $N_\ell$ and a constant $M_0$ such that \begin{equation}\label{ml}M_{N_\ell}\leqslant M_0.\end{equation} To simplify notation, we continue to write $N$ for this subsequence. The spectrum of a Kronecker product consists of products of the factor spectra, and hence $\|\R^{(d)}\|_{\mathrm{op}}=\|\R_{\mathrm{1d}}\|_{\mathrm{op}}^d$ by \eqref{R1d}. Since $\R_{\mathrm{1d}}$ is symmetric, its operator 2-norm is at most its maximum absolute row sum. We have
\begin{equation*}
 \|\R_{\mathrm{1d}}\|_{\mathrm{op}}\leqslant\max_i\sum_{j=1}^n|q^{|i-j|}|=\max_i\sum_{j=1}^nq^{|i-j|}
 \leqslant1+2\sum_{h=1}^{n-1}q^h
 \leqslant1+\frac{2}{1-q}.
\end{equation*}
Along the subsequence satisfying \eqref{ml}, $m_N=M_N^{1/d}\leqslant M_0^{1/d}$, and therefore $q=e^{-1/m_N}\leqslant q_0:=(e^{-1/M_0})^{1/d}<1$. It follows that
\begin{equation}\label{Rnorm}
 \|\R^{(d)}\|_{\mathrm{op}}=\|\R_{\mathrm{1d}}\|_{\mathrm{op}}^d\leqslant L_0:=\left(1+\frac{2}{1-q_0}\right)^d.
\end{equation}

For every $\rho>0$, if $\rho\geqslant \|\R^{(d)}\|_{\mathrm{op}}$, then \eqref{eq:lowerBV} gives \begin{equation*}B_N\geqslant\|\f_N\|_N^2/4;\end{equation*} if $\rho<\|\R^{(d)}\|_{\mathrm{op}}$, then combining \eqref{eq:lowerBV} and \eqref{Rnorm} gives
\begin{equation*}
 V_N\geqslant\frac{\sigma^2}{(2\|\R^{(d)}\|_{\mathrm{op}})^2}
 \geqslant\frac{\sigma^2}{4L_0^2}.
\end{equation*}
Therefore, along this subsequence satisfying \eqref{ml} and for every deterministic $\rho_N$,
\begin{equation}\label{rie}
 \Risk_N(\theta_N,\rho_N;\f_N)\geqslant
 \min\left\{\frac{\|\f_N\|_N^2}{4},
 \frac{\sigma^2}{4L_0^2}\right\}.
\end{equation}Note that continuity of the fixed $f$ and Riemann summation on the bounded domain give
\begin{equation*}
 \|\f_N\|_N^2\longrightarrow \|f\|_{L_2([0,1]^d)}^2>0.
\end{equation*}
The right-hand side of \eqref{rie} has a positive limit inferior, which proves \eqref{eq:positiverisk}.
\end{proof}

\section*{Acknowledgements}

This work is partially supported by the National Natural Science Foundation of China (grant number 12571276).



\begin{thebibliography}{99}
\footnotesize
\raggedright
\setlength{\itemsep}{0.35em}

\bibitem{baker}
Baker, E., Barbillon, P., Fadikar, A., Gramacy, R. B., Herbei, R., Higdon, D., Huang, J., Johnson, L. R., Ma, P., Mondal, A., Pires, B., Sacks, J., \& Sokolov, V. (2022). Analyzing stochastic computer models: A review with opportunities. \textit{Statistical Science}, 37, 64--89.

\bibitem{barthelme2023gaussian}
Barthelm\'e, S., Amblard, P. O., Tremblay, N., \& Usevich, K. (2023). Gaussian process regression in the flat limit. \textit{The Annals of Statistics, 51}(6), 2471--2505.

\bibitem{barthelme2021spectral}
Barthelm\'e, S., \& Usevich, K. (2021). Spectral properties of kernel matrices in the flat limit. \textit{SIAM Journal on Matrix Analysis and Applications, 42}(1), 17--57.

\bibitem{caponnetto2007optimal}
Caponnetto, A., \& De Vito, E. (2007). Optimal rates for the regularized least-squares algorithm. \textit{Foundations of Computational Mathematics, 7}(3), 331--368.

\bibitem{Chung}
Chung, F. R. K. (1997). \textit{Spectral Graph Theory}. American Mathematical Society.

\bibitem{dicker2017kernel}
Dicker, L. H., Foster, D. P., \& Hsu, D. (2017). Kernel ridge vs. principal component regression: Minimax bounds and the qualification of regularization operators. \textit{Electronic Journal of Statistics, 11}(1), 1022--1047.

\bibitem{driscoll2002interpolation}
Driscoll, T. A., \& Fornberg, B. (2002). Interpolation in the limit of increasingly flat radial basis functions. \textit{Computers \& Mathematics with Applications, 43}(3--5), 413--422.


\bibitem{Horn}
Horn, R. A., \& Johnson, C. R. (2012). \textit{Matrix Analysis (2nd ed.)}. Cambridge University Press.

\bibitem{kaufman2013range}
Kaufman, C. G., \& Shaby, B. A. (2013). The role of the range parameter for estimation and prediction in geostatistics. \textit{Biometrika, 100}(2), 473--484.

\bibitem{pramanick}
Pramanick, S., \& Karvonen, T. (2026). Lengthscale estimation for the Gaussian kernel and deterministic response functions. arXiv:2611.13930. \url{https://doi.org/10.48550/arXiv.2611.13930}

\bibitem{rasmussen2006gaussian}
Rasmussen, C. E., \& Williams, C. K. I. (2006). \textit{Gaussian processes for machine learning}. MIT Press.


\bibitem{sanz2026matern}
Sanz-Alonso, D. (2026). \textit{Non-asymptotic analysis of Mat\'ern regression: The roles of target and kernel lengthscales}. arXiv:2608.22553. \url{https://arxiv.org/abs/2608.22553}

\bibitem{stein1999interpolation}
Stein, M. L. (1999). \textit{Interpolation of spatial data: Some theory for kriging}. Springer.

\bibitem{steinwart2009optimal}
Steinwart, I., Hush, D. R., \& Scovel, C. (2009). Optimal rates for regularized least squares regression. In \textit{Proceedings of the 22nd Annual Conference on Learning Theory} (pp. 79--93).

\bibitem{vandervaart2008rates}
van der Vaart, A. W., \& van Zanten, J. H. (2008). Rates of contraction of posterior distributions based on Gaussian process priors. \textit{The Annals of Statistics, 36}(3), 1435--1463.

\bibitem{vandervaart2009adaptive}
van der Vaart, A. W., \& van Zanten, J. H. (2009). Adaptive Bayesian estimation using a Gaussian random field with inverse Gamma bandwidth. \textit{The Annals of Statistics, 37}(5B), 2655--2675.

\bibitem{vandervaart2011information}
van der Vaart, A. W., \& van Zanten, J. H. (2011). Information rates of nonparametric Gaussian process methods. \textit{Journal of Machine Learning Research, 12}, 2095--2119.

\bibitem{wahba1990spline}
Wahba, G. (1990). \textit{Spline models for observational data}. SIAM.

\bibitem{wang2022gaussian}
Wang, W., \& Jing, B.-Y. (2022). Gaussian process regression: Optimality, robustness, and relationship with kernel ridge regression. \textit{Journal of Machine Learning Research, 23}(193), 1--67.

\bibitem{Wendland}Wendland, H. (2005). \textit{Scattered Data Approximation}. Cambridge University Press, Cambridge.

\bibitem{wenzel2026shape}
Wenzel, T., \& Santin, G. (2026). \textit{On the optimal shape parameter for kernel methods: Sharp direct and inverse statements}. arXiv:2601.14070. \url{https://arxiv.org/abs/2601.14070}

\bibitem{zhang2004inconsistency}
Zhang, H. (2004). Inconsistent estimation and asymptotically equal interpolations in model-based geostatistics. \textit{Journal of the American Statistical Association, 99}(465), 250--261.

\end{thebibliography}
\end{document}